\documentclass[sigconf]{acmart}
\usepackage{graphicx}
\usepackage{booktabs}
\usepackage{array}
\usepackage{multirow}
\usepackage{enumitem}
\usepackage{here}
\usepackage{bm}
\usepackage{cleveref}
\usepackage{float}
\usepackage{here}
\usepackage{tabularx}

\acmSubmissionID{50}

\begin{document}

\title{Learning Gaussian Data Augmentation in Feature Space for One-shot Object Detection in Manga}

\author{Takara Taniguchi}
\orcid{0009-0002-0546-2178}
\affiliation{%
  \institution{The University of Tokyo}
  \country{Japan}}
\email{hiroshi-tani@g.ecc.u-tokyo.ac.jp}

\author{Ryosuke Furuta}
\orcid{0000-0003-1441-889X} 
\affiliation{%
  \institution{The University of Tokyo}
  \country{Japan}}
\email{furuta@iis.u-tokyo.ac.jp}


\begin{abstract}
We tackle one-shot object detection in Japanese Manga. The rising global popularity of Japanese manga has made the object detection of character faces increasingly important, with potential applications such as automatic colorization. However, obtaining sufficient data for training conventional object detectors is challenging due to copyright restrictions. Additionally, new characters appear every time a new volume of manga is released, making it impractical to re-train object detectors each time to detect these new characters. Therefore, one-shot object detection, where only a single query (reference) image is required to detect a new character, is an essential task in the manga industry.
One challenge with one-shot object detection in manga is the large variation in the poses and facial expressions of characters in target images, despite having only one query image as a reference. Another challenge is that the frequency of character appearances follows a long-tail distribution. To overcome these challenges, we propose a data augmentation method in feature space to increase the variation of the query. The proposed method augments the feature from the query by adding Gaussian noise, with the noise variance at each channel learned during training. The experimental results show that the proposed method improves the performance for both seen and unseen classes, surpassing data augmentation methods in image space.

\end{abstract}

\ccsdesc[500]{Computing methodologies~Object detection}

  \keywords{One-shot object detection, Manga, Data augmentation}
\maketitle

\newcolumntype{M}[1]{>{\centering\arraybackslash}m{#1}}

\section{Introduction}
\label{sec:intro}

Manga (Japanese comics) are sold and read worldwide, making automatic content recognition essential for the future of the manga market.
Object detection in manga is one of the most important tasks because of its wide variety of applications.
For example, object detection approaches have been applied to text detection and segmentation in manga~\cite{mangadetection_text_1_chu,mangadetection_text_2_aramaki,mangadetection_del_text,mangadetection_text_piriyothinkul} for the purposes of automatic translation.
Another example is face detection\cite{mangadetection_face_qin, aukkapinyo2023manga}, which can be applied to automatic colorization or speaker estimation in manga.

Despite the wide range of practical applications, obtaining sufficient data to train conventional object detectors is difficult due to copyright issues with manga.
In addition, when the detection target is a character's face or body, this problem becomes more serious because new characters appear every time a new volume is released.
In this case, it is not practical to re-train conventional object detectors each time to detect these new characters because the target classes must be fixed between training and testing.
Open vocabulary object detection has been actively researched in recent years~\cite{zareian2021open,minderer2022simple,li2023gligen}, but methods in this vein can only handle general object class names and cannot address newly introduced character names.

\begin{figure}[t]
    \centering
    \includegraphics[width=1\linewidth]{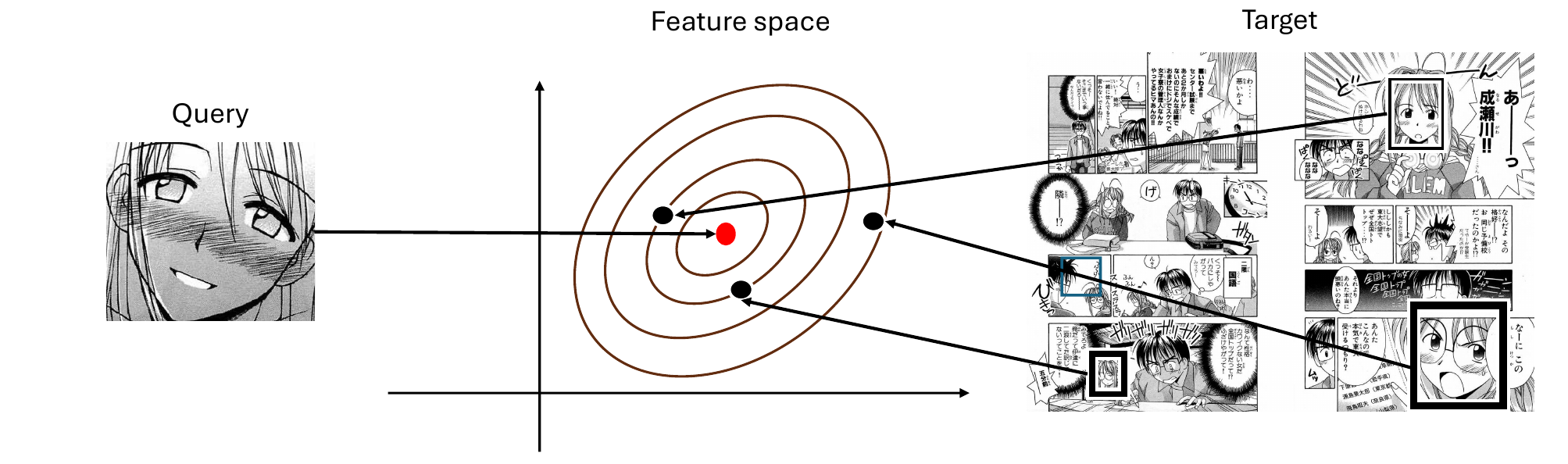}
    \caption{Key idea of our method. In one-shot object detection in manga, there is a large variation in the poses and facial expressions of characters in target images, despite having only one query image as a reference. We propose a Gaussian noise-based data augmentation in the feature space, assuming that feature vectors from the same class are normally distributed in the feature space even though they exhibit large variations in the image space.}
    \label{fig:intro_overview}
\end{figure}

In this paper, we tackle the task of one-shot object detection in manga, where not only seen but also unseen classes during training (e.g., new characters) can be detected using a single query (reference) image.
A major challenge of one-shot object detection in manga is the large variation in the poses and facial expressions of characters in the target images, despite having only one query image as a reference, as shown in Fig.~\ref{fig:intro_overview}.
Another challenge is that the frequency of character appearances follows a long-tail distribution.
For example, main characters tend to appear frequently while supporting characters may appear only once or twice.
In addition, it is difficult to collect large amounts of training data due to copyright issues, as mentioned above.
To overcome these challenges, we propose a data augmentation method in feature space to increase the variation of the query.
The key idea of our method is the assumption that feature vectors from the same class are normally distributed in the feature space, even though they exhibit large variations in the image space.
On the basis of this assumption, we augment the feature from the query by adding Gaussian noise.
Since different channels in feature maps can capture different patterns, the noise variance in each channel is learned during the training.

Our experimental results show that incorporating the proposed method into a state-of-the-art one-shot object detector improves the performance for both seen and unseen classes.
Also, we demonstrate that the proposed method outperforms commonly utilized data augmentation methods in image space.

Our contributions are summarized as follows:
\begin{enumerate}
    \item[-] We tackle one-shot object detection in Japanese manga for the first time.
    \item[-] We propose a data augmentation method in feature space based on Gaussian noise with learnable variances to address the challenges of large variation and long-tailed distributions of manga characters.
    \item[-] We demonstrate that the proposed method improves the performance of a state-of-the-art method for one-shot detection on the Manga109 dataset.
\end{enumerate}

\section{Related Works}
\subsection{Object detection in manga}
Existing studies on object detection in manga have mainly focused on conventional object detection, not one-shot detection. 
While one-shot object detection can detect classes that only appear in the testing phase as well as classes used for training by utilizing a query image, conventional object detection methods are limited to detecting classes that are included in their training data. 
Within the context of object detection in manga, classes such as face~\cite{mangadetection_face_qin, aukkapinyo2023manga}, text~\cite{mangadetection_text_1_chu,mangadetection_text_2_aramaki,mangadetection_del_text,mangadetection_text_piriyothinkul}, frame\cite{mangadetection_zhou}, and body~\cite{yanagisawa2018study} have been studied.
Aukkapinyo \textit{et al.}~\cite{aukkapinyo2023manga} particularly focused on the drawing styles of characters’ faces in manga, while Zhou \textit{et al.}~\cite{mangadetection_zhou} tackled the frame detection task by using heuristics and a CNN-based model. 
Yanagisawa \textit{et al.}~\cite{yanagisawa2018study} investigated object detection tasks for four classes appearing in the Manga109\cite{manga109} dataset.
Ogawa \textit{et al.}~\cite{mangadetection_ogawa} tackled the problem attributed to anchor-based detectors in Japanese Manga.
However, all of these studies are limited to conventional object detection tasks where target classes are fixed between training and testing. 
In contrast, our work tackles one-shot object detection in manga for the first time.

\subsection{One-shot object detection}
One-shot object detection~\cite{oneshot_hsieh,oneshot_michaelis_oneseg,oneshot_chen_ait,oneshot_yang_focosod,oneshot_xiang_li, oneshot_oscd_kun_fu,oneshot_anton} is a specialized form of object detection, where the objective is to detect objects of the specified class in the target image based on a single query image. This technique is capable of detecting both seen classes, i.e., classes that were utilized during the training phase, and unseen classes, i.e., those that were not included in the training set.

Michaelis \textit{et al.}~\cite{oneshot_michaelis_oneseg} addressed feature matching for the one-shot object detection task.
Hsieh \textit{et al.}~\cite{oneshot_hsieh} explored the use of a non-local proposal, co-excitation, and proposal ranking scheme.
Yang \textit{et al.}~\cite{oneshot_yang_focosod} reduced the number of false positive samples commonly observed in one-shot object detection tasks by concentrating on the classification task.
Chen \textit{et al.}~\cite{oneshot_chen_ait} implemented an attention-based encoder-decoder architecture for one-shot object detection. 
Our work differs from these in that we address one-shot object detection tasks specifically for Japanese manga images.

\subsection{Data augmentation}
There are numerous data augmentation methods that can be directly applied to images\cite{data_aug_cutout_devries,data_aug_moreuno-barea_noiseinjection, image_aug_radon_nanni, data_aug_chatfield_color,data_aug_inoue, data_aug_zhong}.
For example, Nanni \textit{et al.}~\cite{image_aug_radon_nanni} introduced the Radon transform and Fourier transform to augment data. 
Moreno-Barea \textit{et al.}~\cite{data_aug_moreuno-barea_noiseinjection} introduced noise to achieve a variety of representations.
Conversely, there are data augmentation methods applied to the feature space\cite{feature_aug_lee,aug_feature_chen,feature_aug_gaussian_li,feature_aug_jia, data_aug_liu, aug_data_gastaldi_shake,data_aug_shen_feature}.
For example, Shen \textit{et al.}~\cite{data_aug_shen_feature} applied affine transformations such as rotation and scale to feature maps. 
In addition, there are GAN-based~\cite{GAN_original} data augmentation approaches\cite{data_aug_gan_frid-adar, aug_auggan_frid-adar,gan_aug_tanaka}. Moreover, Yang and Chu \cite{mangadetection_aug_tet_yang} developed an effective data augmentation method using a GAN\cite{GAN_original} specifically designed for manga images.
In contrast, we propose a simple yet effective data augmentation method to address the challenges of large variation and long-tail distribution in one-shot detection in manga. In our method, a query feature is augmented by adding Gaussian noise. The variance of the noise is trained using the detection loss.

\section{Proposed Method}
\label{sec:methods}

\subsection{Problem definition}
In the task of one-shot object detection, we have a set of seen classes $\mathcal{S}$ and a set of unseen classes $\mathcal{U}$. 
They have no intersection (i.e., $\mathcal{S} \cap \mathcal{U} = \phi$).
To detect these classes, the one-shot object detector requires a query patch $q$. It is trained to detect all instances corresponding to the query in the target image $t$. In the training process, only the set of seen classes $\mathcal{S}$ is used. Once trained, it can detect unseen classes $\mathcal{U}$ in the inference process.

\subsection{Overall architecture}

\begin{figure*}[t]
    \centering
    \includegraphics[width=0.8\linewidth]{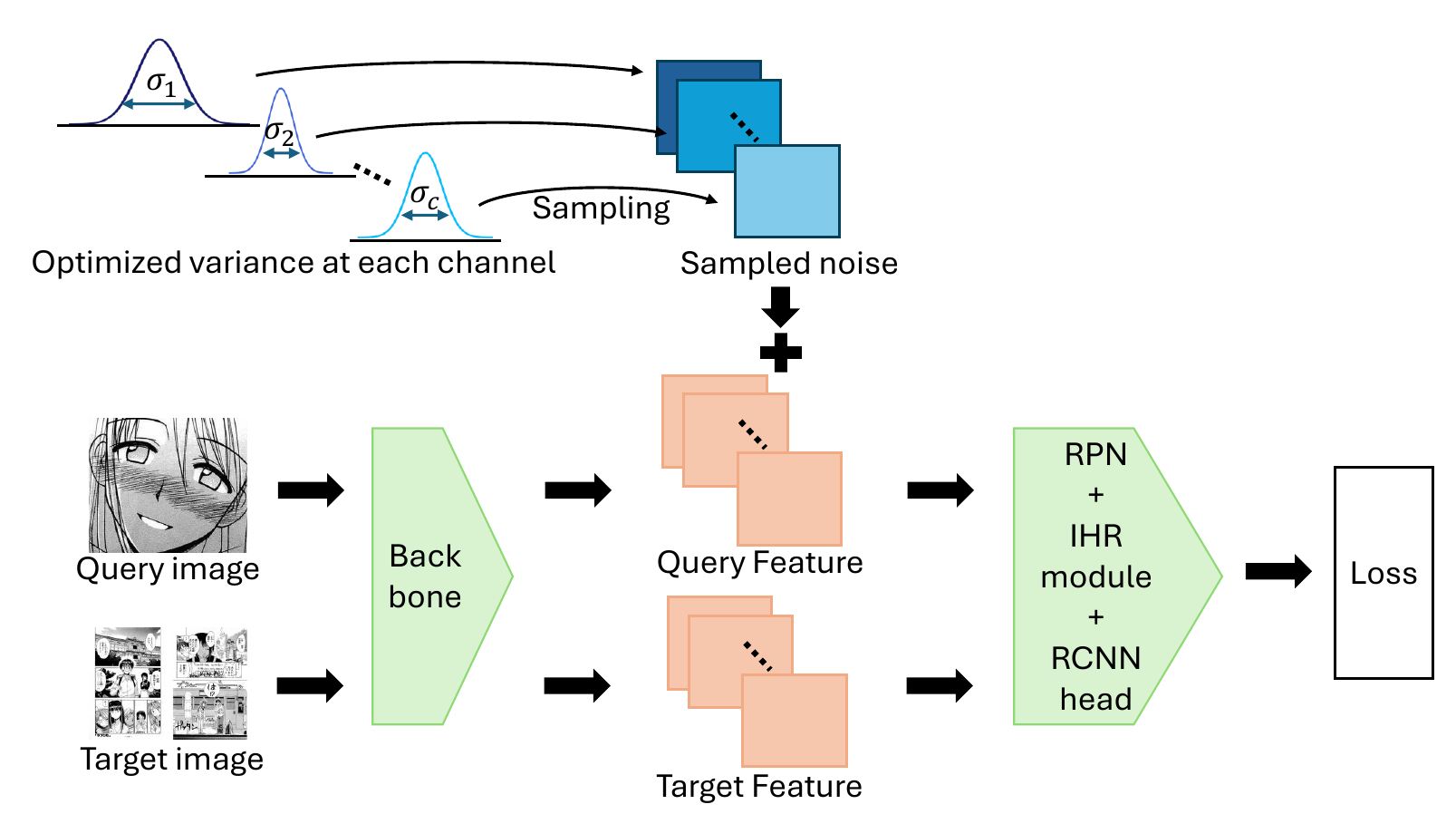}
    \caption{Overview of the proposed method. After extracting feature maps from the query and target images, Gaussian noise is added to the query feature to augment it according to the Gaussian distribution. Then, the target feature and the augmented query feature are fed into the RPN, IHR module, and RCNN head to obtain detection results, similar to BHRL~\cite{BHRL}. The variance of the Gaussian noise is optimized at each channel during the training.}
    \label{model_architecture}
\end{figure*}

Fig. \ref{model_architecture} presents an overview of our framework. Similar to other one-shot object detection models, our detector takes a pair of query and target images as input.
Our detector is based on BHRL~\cite{BHRL}, which is a state-of-the-art method for one-shot object detection.
We denote the parameters of the detection network as $\bm{\theta}$.

Given a pair of a query image $q$ and a target image $t$ as input, we first extract a query feature $F^q\in \mathbb{R}^{C\times H\times W}$ and a target feature $F^t\in \mathbb{R}^{C\times H\times W}$ through a backbone network. 
$C, H$, and $W$ are the number of channels, height, and width of the feature maps, respectively.
To augment the query feature $F^q$ during the training, we sample noise $N\in\mathbb{R}^{C\times H\times W}$ from a Gaussian distribution and add it to the query feature (details explained in Sec.~\ref{sec:noise}).
After that, we follow the same procedure as BHRL~\cite{BHRL}.
Specifically, we feed the query and target features into the matching module~\cite{michaelis2018one} and the Region Proposal Network (RPN) to generate a set of object proposals.
Using each proposal, we pool the instance-level target feature $F'^t\in\mathbb{R}^{C'\times K\times K}$ from $F^t$ with the RoI pooling operation, where $C'$ and $K$ are the number of channels and size of the instance-level features.
From $F^q$, we pool the instance-level query feature $F'^q\in\mathbb{R}^{C'\times K\times K}$ using the input query patch region with the RoI pooling.
Next, the two instance-level features are fused by the Instance-level Hierarchical Relation (IHR) module proposed in~\cite{BHRL}, which considers three types of relations between $F'^q$ and $F'^t$: contrastive relation, salient relation, and attention relation.
Finally, the fused feature is fed into the R-CNN head to obtain the detection results (i.e., bounding boxes and class scores).
At the training phase, we compute the loss $\mathcal{L}$, which consists of the L1 loss and the Ratio-Preserving Loss proposed in~\cite{BHRL}, and update the network parameters as
\begin{equation}
\bm{\theta}\leftarrow \bm{\theta} - \gamma \frac{\partial \mathcal{L}}{\partial \bm{\theta}},
\end{equation}
where $\gamma$ is the learning rate.

\subsection{Learning Gaussian data augmentation}\label{sec:noise}

\begin{figure}[t]
    \centering
    \includegraphics[width=0.7\linewidth]{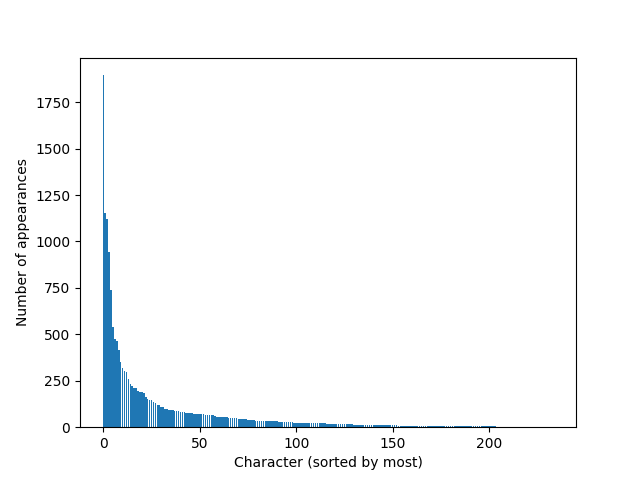}
    \caption{Distribution of the number of appearances for each character.}
    \label{fig:distribution_character}
\end{figure}

One major challenge with one-shot object detection in manga is the large variation in poses and facial expressions of characters in target images, despite having only one query image as a reference, as shown in Fig.~\ref{fig:intro_overview}.
Another challenge is that the frequency of character appearances follows a long-tail distribution as shown in Fig.~\ref{fig:distribution_character}.
In addition, it is difficult to collect large amounts of training data due to copyright issues.
A common approach to overcome these challenges is the implementation of data augmentation to increase the variation of input images.
However, typical data augmentation approaches on image space (e.g., random flipping, cropping, and color jittering) cannot deal with these variations.
Therefore, we propose a Gaussian noise-based data augmentation in the feature space, assuming that feature vectors from the same class are normally distributed in the feature space even though they exhibit large variations in the image space.

To augment the query feature $F^q$ according to the normal distribution, we sample a Gaussian noise $N=[n_{ijk}]_{C\times H\times W}$ (where $i=1,\cdots, C$ and $j=1,\cdots, H$ and $k=1,\cdots, W$) and add it to $F^q$.
Since different channels of the feature maps capture different semantics, we utilize different normal distributions (same zero mean, but different standard deviations $\sigma_i$) at different channels $(i=1,\cdots, C)$.
Specifically, the proposed data augmentation is performed as follows:
\begin{gather}
n_{ijk}\sim \mathcal{N}(0,\sigma^2_i),\label{eq:sample}\\
F^q \leftarrow F^q + N.
\end{gather}

Since it is difficult to manually tune $\sigma_i$ for each channel, we optimize $\sigma_i$ from data during the training.
To achieve this, we adopt differentiable sampling using the reparameterization trick~\cite{kingmaauto} instead of Eq. (\ref{eq:sample}), as
\begin{equation}
    n_{ijk}=\sigma_i \epsilon, \ \ \epsilon \sim \mathcal{N}(0, 1).
\end{equation}
The standard deviations $\bm{\sigma}=(\sigma_1,\cdots,\sigma_C)$ are optimized simultaneously with the network parameters $\bm{\theta}$ by minimizing the same loss:
\begin{equation}
\bm{\sigma}\leftarrow \bm{\sigma} - \gamma \frac{\partial \mathcal{L}}{\partial \bm{\sigma}}.\label{eq:backpropagation}
\end{equation}

Note that the proposed data augmentation is performed only at the training phase.
Therefore, there are no extra computational costs at the inference phase.

\section{Experiment}
\label{sec:exp}

\subsection{Settings}
\subsubsection{Datasets}
We conducted the experiments on the Manga109\cite{manga109} dataset. 
This dataset is composed of a variety of 109 Japanese comics and consists of annotations of four classes: frames, texts, faces, and bodies.
We used characters' name tags as class labels and their face bounding boxes as regression targets.
For the one-shot setting, we chose the five manga titles that have two volumes in the Manga109 dataset, as shown in Table~\ref{tab:manga_comparison}.
We used the earlier volumes of the five titles for training and the rest of the five volumes for testing.
In this setting, we can naturally define the seen and unseen classes because some characters appear in both the earlier and later volumes, while other characters appear only in the later volumes.
The numbers of seen and unseen classes are 137 and 97 in total, respectively.

\begin{table}[t]
\centering
\caption{Manga volumes used in our experiments.}
\label{tab:manga_comparison}
\begin{tabular}{@{}clcccc@{}}
\toprule
Split & Title & Volume & \#Pages & \#Images \\ \midrule
\multirow{5}{*}{Train} & HighschoolKimengumi & 1 & 197 & 99 \\
& LoveHina & 1 & 192 & 97 \\
& Moeruonisan & 1 & 183 & 92 \\
& Saladdays & 1 & 182 & 91 \\
& ShimatteIkouze & 1 & 215 & 108 \\ \midrule
\multirow{5}{*}{Test} & HighschoolKimengumi & 20 & 187 & 94 \\
& LoveHina & 14 & 194 & 97 \\
& Moeruonisan & 19 & 197 & 99 \\
& Saladdays & 18 & 190 & 95 \\
& ShimatteIkouze & 26 & 223 & 112 \\
\bottomrule
\end{tabular}
\end{table}

\subsubsection{Implementation details}
We implemented the proposed method based on the publicly available code of BHRL~\cite{BHRL}, which is built on the MMDetection library~\cite{mmdetection} using the PyTorch framework~\cite{paszke2019pytorch}. 
Similar to BHRL, we used ResNet-50~\cite{resnet50} with FPN~\cite{lin2017feature} as the backbone network.
The ResNet-50 network was initialized with the ImageNet pre-trained model, and the other parameters were initialized randomly.
The standard deviations $\sigma_i$ were initialized as 0.1.
We trained the proposed network for 30 epochs, which took 2 hours on an NVIDIA Tesla A100 GPU.
All the hyperparameters were the same as BHRL except for the number of training epochs.

\subsubsection{Evaluation metrics}

We followed the evaluation metrics in BHRL\cite{BHRL} and used $AP_{50}$ for evaluation.
However, in contrast to general one-shot object detection datasets such as PASCAL VOC and MSCOCO, Manga109~\cite{manga109} contains a large number of characters who rarely appear, as shown in Fig. \ref{fig:distribution_character}.
If we simply average $AP_{50}$ on all character classes, the detection performance on those rare classes would have a significant impact on the score.
To evaluate methods based on the number of characters' appearances, we set thresholds for the number of appearances of the characters included in the inference process.
Specifically, $thr = X$ means that we ignore the character classes that appear less than $X$ times in the test set.
In other words, when $thr$ is set to a high value (e.g., 320), only the characters that appeared many times (e.g., 320 or more times) were included in the evaluation.
$thr = 0$ means all that the character classes were used for the evaluation.

\subsection{Comparisons with other augmentation methods}

\subsubsection{Baselines}
As there are no existing data augmentation methods in feature space that can be directly applied to one-shot object detection, we compare the proposed method with other data augmentation methods in image space that are utilized in object detection~\cite{chen2022learning}: Gaussian blur (Gblur), solarize, and random crop (Rcrop).
Note that four simple data augmentations (resize, flip, normalize, and pad) are performed by default in BHRL.
We performed the proposed and baseline augmentation methods in addition to those four default augmentations.
All the augmentation is applied to query images for fair comparisons.
For the random crop, the crop size is set to (64,64).
In Gaussian blur, the kernel size is set to 3, with its variance chosen uniformly at random between 0.1 to 2.0 at each time. 
In the solarize operation, it is randomly applied with a probability of 0.5, and its inverting threshold is set to 100.

\subsubsection{Results}

\begin{table*}[t]
\centering
\caption{Comparisons of $AP_{50}$ between the proposed and other data augmentation methods based on the number of characters' appearances in the test dataset.
}
\label{augmentation_compare}
\setlength{\tabcolsep}{15pt} 
\begin{tabular}{@{}crrrrrrrrr@{}}
\toprule
         & \multicolumn{4}{c}{seen}      & \multicolumn{5}{c}{unseen}            \\ \cmidrule(l{2pt}r{2pt}){2-5} \cmidrule(l{2pt}r{2pt}){6-10}
thr      & 0     & 80    & 160   & 320   & 0     & 20    & 40    & 80    & 100   \\
\midrule
Default   & 0.183 & 0.201 & 0.259 & 0.319 & 0.240  & 0.241 & 0.230  & 0.309 & 0.316 \\
+Ours       & \bf{0.206} & \bf{0.219} & \bf{0.275} & \bf{0.322} & \bf{0.260} & 0.251 & \bf{0.241} & \bf{0.322} & \bf{0.347} \\
+Gblur    & 0.175 & 0.192 & 0.236 & 0.286 & 0.241 & 0.256 & 0.233 & 0.318 & 0.336 \\
+Solarize & 0.148 & 0.157 & 0.181 & 0.262 & 0.238 & \bf{0.261} & 0.238 & 0.317 & 0.327 \\
+Rcrop    & 0.125 & 0.142 & 0.192 & 0.289 & 0.183 & 0.200 & 0.181 & 0.256 & 0.269 \\
 \bottomrule
\end{tabular}
\end{table*}

In Table~\ref{augmentation_compare}, we compare the proposed method with other augmentation methods on seen and unseen classes. 
Overall, our method shows an improvement over the default augmentation on both seen and unseen classes. 
This improvement is observed consistently with all the thresholds.
In contrast, GBlur and Solarize decreased the performance on seen classes although they improved on unseen classes with some thresholds.
Rcrop degraded the performance on both seen and unseen classes with all the thresholds.
These results demonstrate that the proposed augmentation method is compatible with the default augmentations in image space, while the baseline methods are not. 
They also indicate that the proposed method, which learns Gaussian noise for data augmentation at each channel to deal with diverse facial expressions and feature space augmentation, is beneficial for the one-shot object detection task in manga images.


\begin{figure*}[t]
    \centering
    \scalebox{1.0}{
    \setlength{\extrarowheight}{60pt} 
    \begin{tabular}{cccc}
        \begin{minipage}{.2\linewidth}
            \centering
            \includegraphics[width=.8\linewidth]{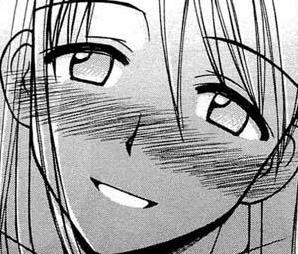}
            Query
        \end{minipage} &
        \begin{minipage}{.2\linewidth}
            \centering
            \includegraphics[width=\linewidth]{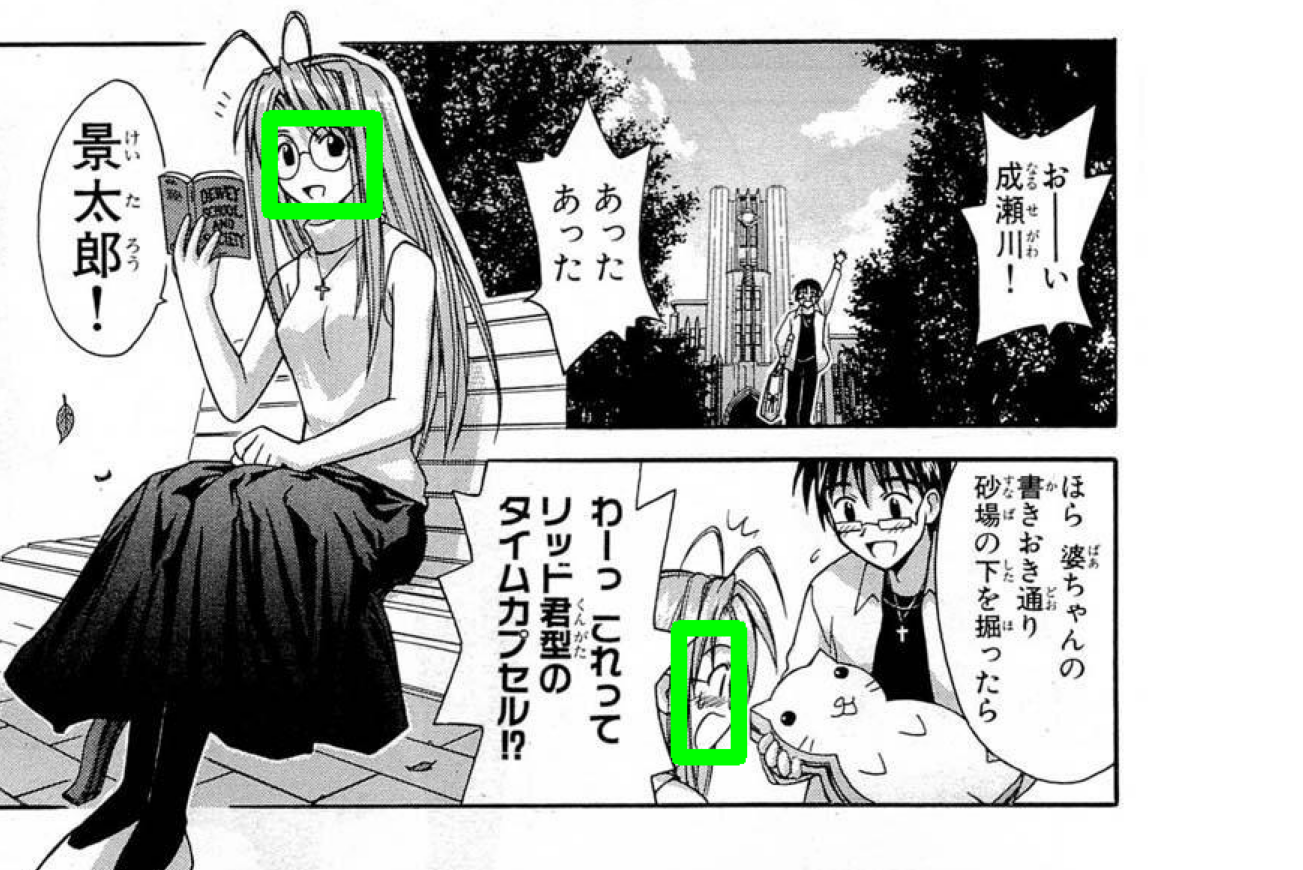}
            GT
        \end{minipage} &
        \begin{minipage}{.2\linewidth}
            \centering
            \includegraphics[width=\linewidth]{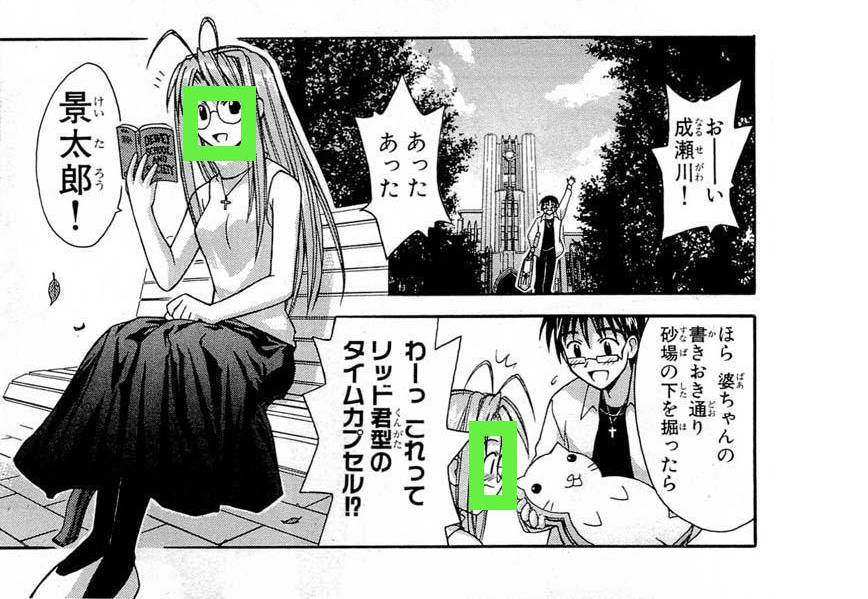}
            Ours
        \end{minipage} &
        \begin{minipage}{.2\linewidth}
            \centering
            \includegraphics[width=\linewidth]{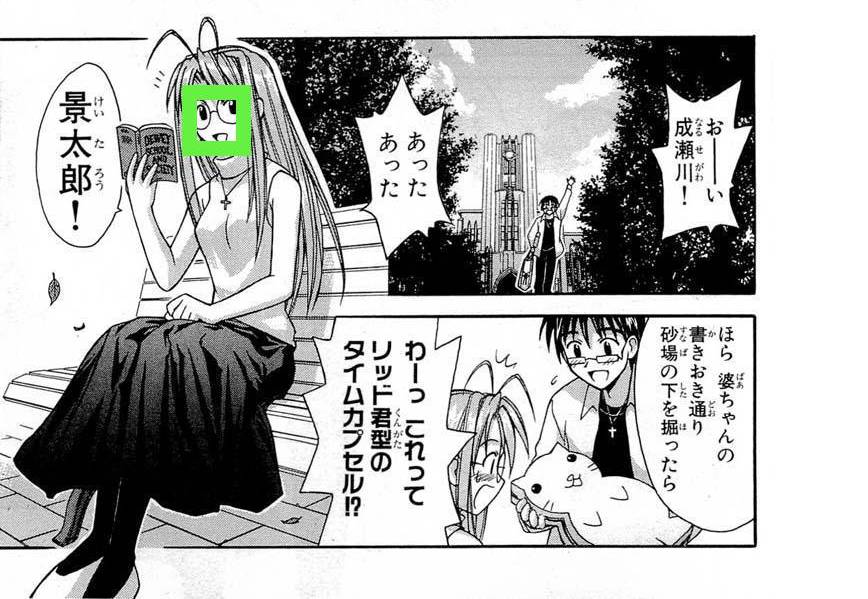}
            Gblur
        \end{minipage} \\ & 
        \begin{minipage}{.2\linewidth}
            \centering
            \includegraphics[width=\linewidth]{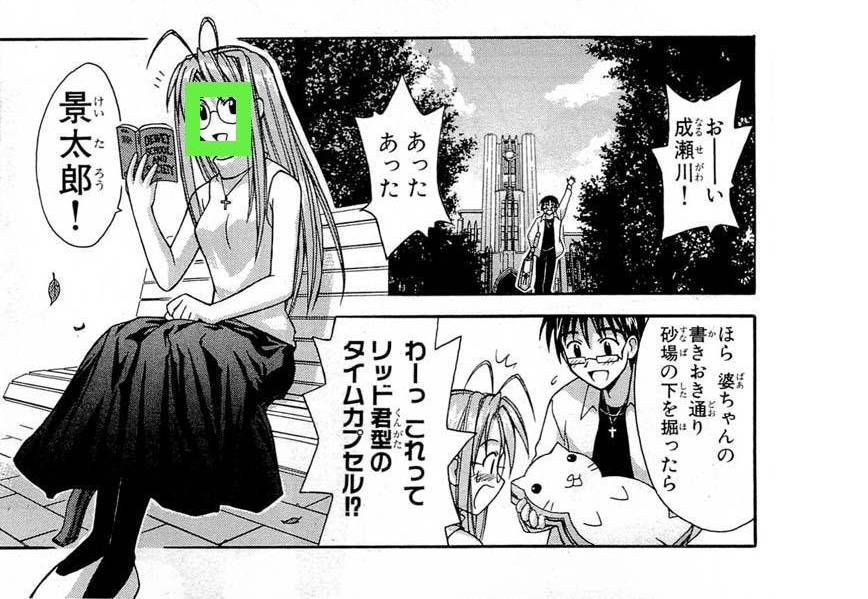}
            Solarize
        \end{minipage} &
        \begin{minipage}{.2\linewidth}
            \centering
            \includegraphics[width=\linewidth]{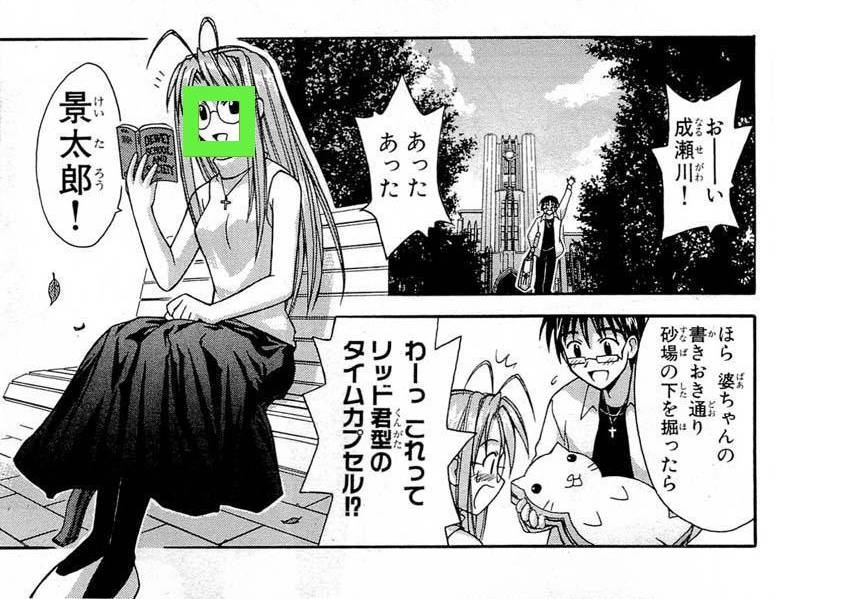}
            Rcrop
        \end{minipage} &
        \begin{minipage}{.2\linewidth}
            \centering
            \includegraphics[width=\linewidth]{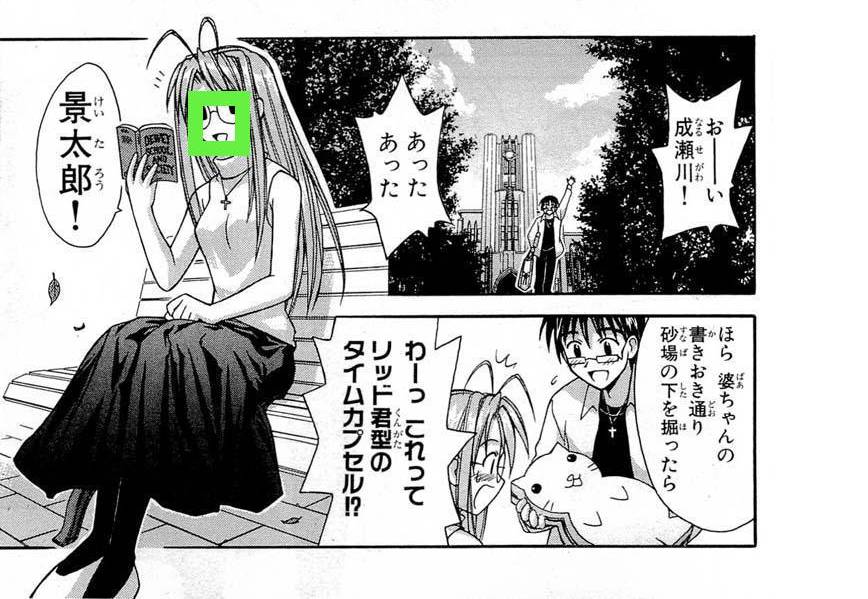}
            Default
        \end{minipage} \\
    \end{tabular}
    }
    \caption{Qualitative comparisons between the proposed and other augmentation methods for seen classes (thr = 320).}
    \label{fig:seen_example}
\end{figure*}

\begin{figure*}[t]
    \centering
    \scalebox{1.0}{
    \setlength{\extrarowheight}{60pt} 

    \begin{tabular}{cccc}
        \begin{minipage}{.2\linewidth}
            \centering
            \includegraphics[width=.8\linewidth]{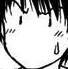}
            Query
        \end{minipage} &
        \begin{minipage}{.2\linewidth}
            \centering
            \includegraphics[width=\linewidth]{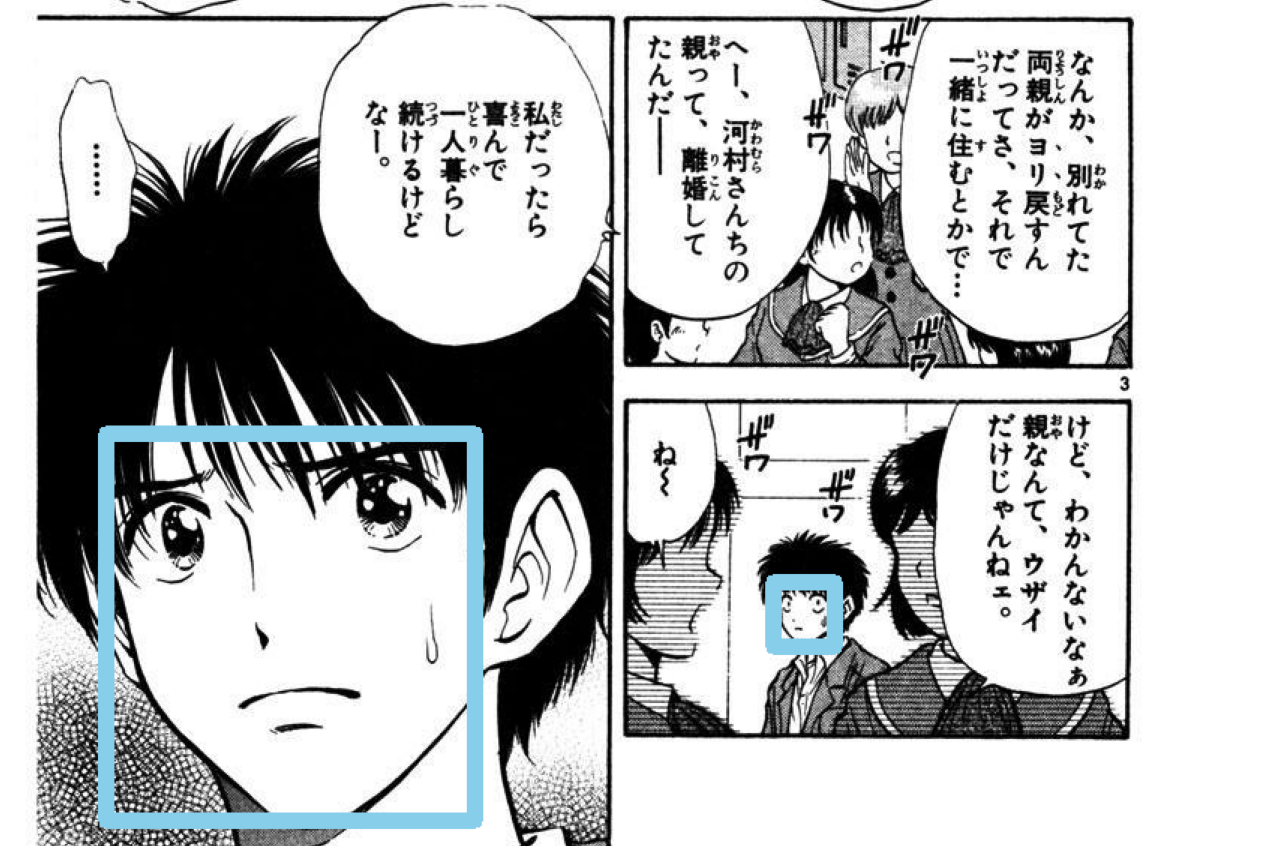}
            GT
        \end{minipage} &
        \begin{minipage}{.2\linewidth}
            \centering
            \includegraphics[width=\linewidth]{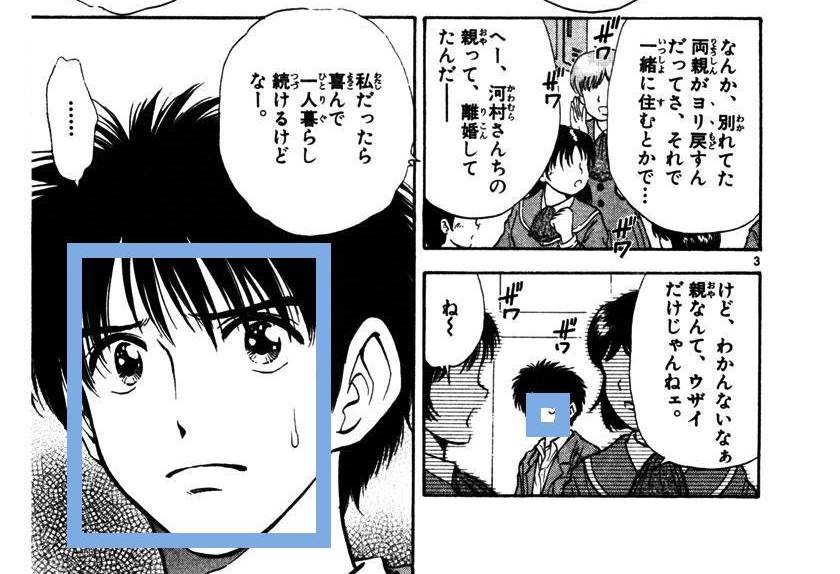}
            Ours
        \end{minipage} &
        \begin{minipage}{.2\linewidth}
            \centering
            \includegraphics[width=\linewidth]{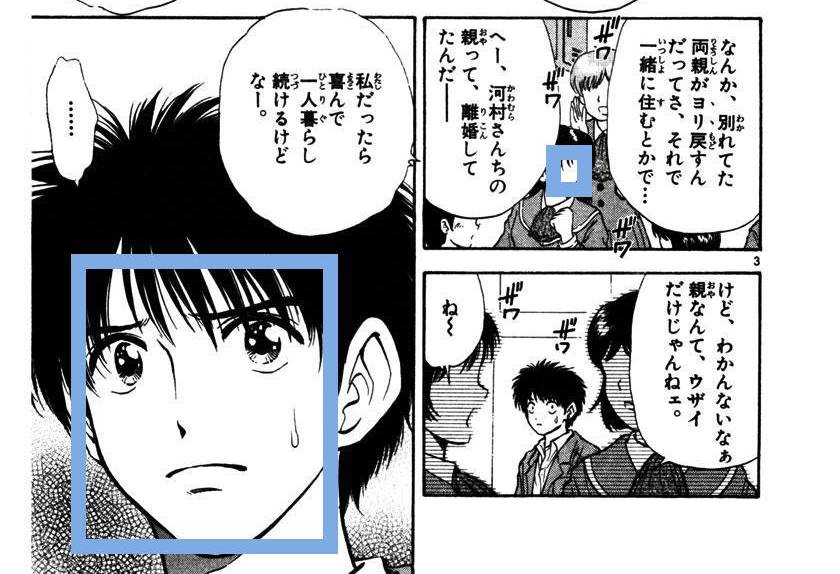}
            Gblur
        \end{minipage} \\ & 
        \begin{minipage}{.2\linewidth}
            \centering
            \includegraphics[width=\linewidth]{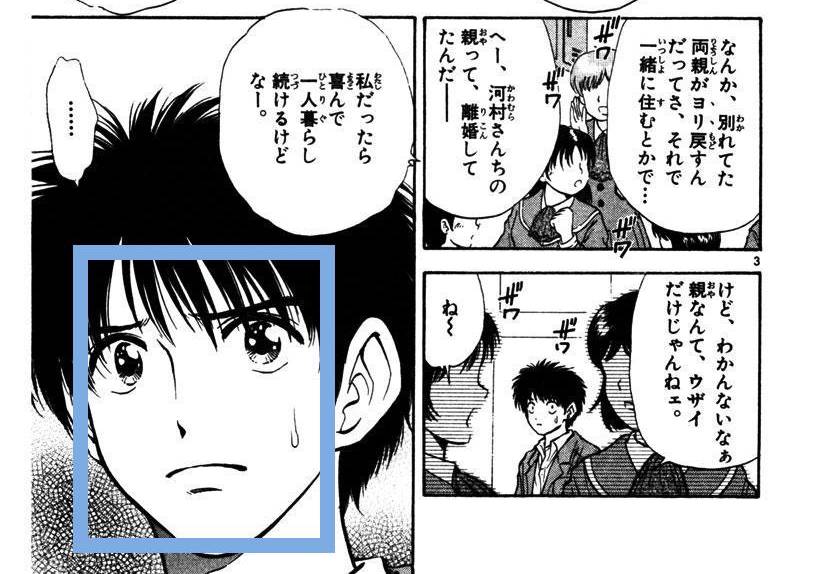}
            Solarize
        \end{minipage} &
        \begin{minipage}{.2\linewidth}
            \centering
            \includegraphics[width=\linewidth]{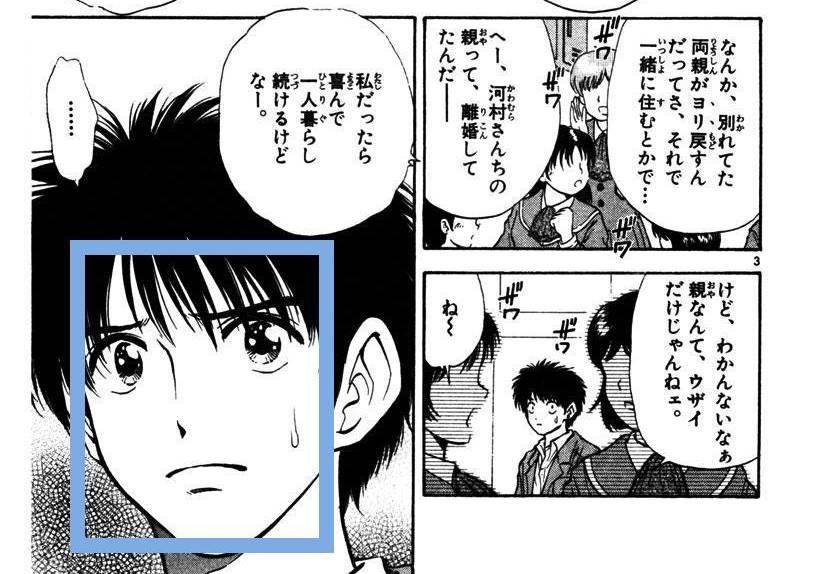}
            Rcrop
        \end{minipage} &
        \begin{minipage}{.2\linewidth}
            \centering
            \includegraphics[width=\linewidth]{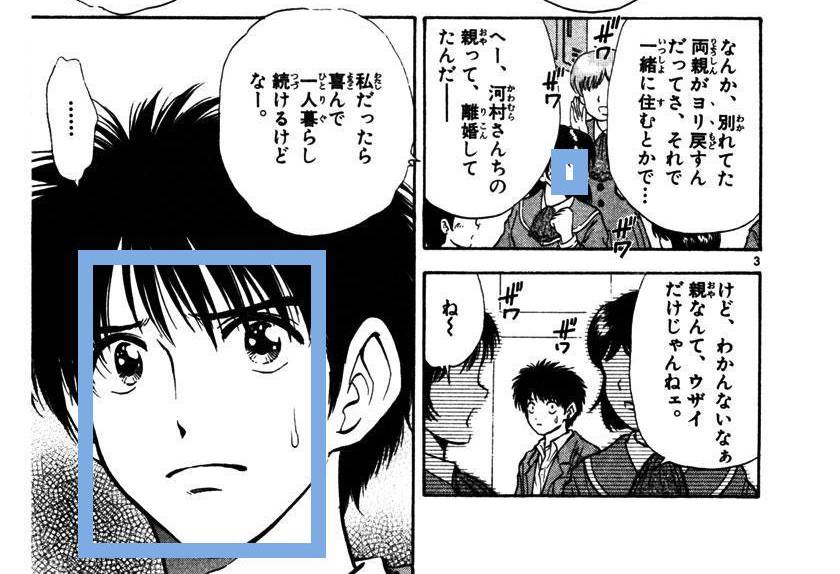}
            Default
        \end{minipage} \\
    \end{tabular}
    }
    \caption{Qualitative comparisons between the proposed and other augmentation methods for unseen classes (thr =100).}
    \label{fig:unseen_example}
\end{figure*}



Fig.~\ref{fig:seen_example} shows the qualitative comparison on a seen class.
In contrast to the proposed method, the baseline methods failed to detect the face on the right because it has a significantly different facial pose from the query.
The qualitative comparison on a unseen class is shown in Fig.~\ref{fig:unseen_example}.
Here, the face on the right has a different facial expression from the query, and therefore, the baseline methods could not detect it correctly.
These results indicate that the proposed method can effectively deal with challenges such as diverse poses and facial expressions by learning the data augmentation in feature space based on Gaussian distributions, while the data augmentation methods in image space cannot.



\subsection{Analysis of learnable variances}

In the proposed method, the variance $\sigma_i^2$ is learned at each channel ($i=1,\cdots, C$) based on the hypothesis that different channels capture different semantics.
In this section, we analyze the effect of the channel-wise learnable variances.
Specifically, we compare the following different types of learnable variances.

\begin{description}
    \item[-] \textbf{Fixed}: The variance is fixed to the initial value 0.1 during the training.
    \item[-] \textbf{Single}: The single variance (shared between channels and positions) is learned during the training.
    \item[-] \textbf{Channel-wise}: The variance is learned at each channel as described in Sec.~\ref{sec:noise}.
    \item[-] \textbf{Position-wise}: The variance is learned at each coordinate (position) in the feature maps.
    \item[-] \textbf{Position-Channel} The variance is learned at each coordinate and channel.
\end{description}

In the Fixed method, the noise sampling is performed as follows:
\begin{equation}
    n_{ijk} \sim \mathcal{N}(0,\sigma^2),
\end{equation}
where the std $\sigma$ is fixed to 0.1 during the training.

In the Single method, the noise is sampled via the differential sampling instead of $n_{ijk}\sim \mathcal{N}(0,\sigma)$, as follows:
\begin{equation}
    n_{ijk}=\sigma \epsilon, \ \ \epsilon \sim \mathcal{N}(0, 1),
\end{equation}
where the $\sigma$ is shared between different channels (and positions) and optimized during the training by Eq.(\ref{eq:backpropagation}).

In the Position-wise method, we have different variances $\sigma_{jk}$ at different positions ($j=1,\cdots,H$ and $k=1,\cdots,W$), but they are shared between different channels.
The noise is sampled via the differential sampling instead of $n_{ijk}\sim \mathcal{N}(0,\sigma_{jk})$, as follows:
\begin{equation}
    n_{ijk}=\sigma_{jk} \epsilon, \ \ \epsilon \sim \mathcal{N}(0, 1),
\end{equation}
where the $\bm{\sigma}=(\sigma_{11},\cdots,\sigma_{HW})$ is optimized during the training by Eq. (\ref{eq:backpropagation}).

In the Position-Channel method, we have different variances $\sigma_{ijk}$ at different positions and channels.
The noise is sampled via the differential sampling instead of $n_{ijk}\sim \mathcal{N}(0,\sigma_{ijk})$ as follows:
\begin{equation}
    n_{ijk}=\sigma_{ijk} \epsilon, \ \ \epsilon \sim \mathcal{N}(0, 1),
\end{equation}
where the $\bm{\sigma}=(\sigma_{111},\cdots,\sigma_{CHW})$ is optimized during the training by Eq. (\ref{eq:backpropagation}).

\subsubsection{Results}

\begin{table*}[h]
\centering
\caption{Comparisons of $AP_{50}$ between different types of learnable variances.}
\label{ablation study}
\setlength{\tabcolsep}{13pt} 
\begin{tabular}{@{}crrrrrrrrrr@{}}
\toprule
       & \multicolumn{4}{c}{seen}      & \multicolumn{5}{c}{unseen}            \\ \cmidrule(l{2pt}r{2pt}){2-5} \cmidrule(l{2pt}r{2pt}){6-10}
thr    & 0     & 80    & 160   & 320   & 0     & 20    & 40    & 80    & 100   \\ \midrule
Default & 0.183 & 0.201 & 0.259 & 0.319 & 0.240  & 0.241 & 0.23  & 0.309 & 0.316 \\
+Fixed  & 0.151 & 0.180 & 0.211 & 0.299 & 0.229 & 0.261 & 0.233 & 0.319 & 0.335 \\
+Single     & 0.130 & 0.137 & 0.165 & 0.240 & 0.213 & 0.258 & 0.229 & 0.304 & 0.315 \\
+Channel-wise     & \bf{0.206} & \bf{0.219} & \bf{0.275} & \bf{0.322} & \bf{0.260} & 0.251 & 0.241 & 0.322 & 0.347 \\
+Position-wise     & 0.184 & 0.193 & 0.235 & 0.274 & 0.253 & 0.261 & 0.257 & 0.336 & \bf{0.354} \\
+Position-Channel   & 0.173 & 0.178 & 0.210 & 0.244 & 0.257 & \bf{0.271} & \bf{0.258} & \bf{0.337} & 0.345 \\
\bottomrule
\end{tabular}
\end{table*}

Table~\ref{ablation study} shows the quantitative comparisons between different learnable variances.
We observe that the Fixed and Single methods degraded the performance on seen classes at all thresholds, although they slightly improved the performance on unseen classes at some thresholds.
In contrast, the Channel-wise method consistently improved the performance on both seen and unseen classes at all the thresholds.
These results indicate that optimizing the variance at each channel is important to diversify the query features for dealing with the large variation of the poses and facial expressions.
The Position-wise and Position-Channel methods significantly improved the performance on unseen classes, but the performance was decreased on seen classes.
We believe this is because having one variance at each position is over-parameterized, making it difficult to optimize with a small amount of training data.

\section{Conclusion}
In this paper, we tackled the new task of one-shot object detection in manga.
To deal with the challenges of this task, namely, the large variations of poses and facial expressions of the characters, the long-tailed distributions of the character classes, and the small amount of training data, we proposed a data augmentation method in feature space based on Gaussian noise.
The experimental results demonstrated that the proposed method improved the performance on both seen and unseen classes by optimizing the variance of the Gaussian noise at each channel.
We also found that the proposed method is compatible with the commonly utilized data augmentation methods in image space.

The limitation is that it is not clear whether the proposed method would still work well when we have a large amount of training data.
In future work, we will investigate the relationship between the performance of the proposed method and the amount of training data.
Also, we will extend the proposed method for test-time data augmentation as well, since augmenting the query image at test time would be beneficial for dealing with the large variation of poses and facial expressions in one-shot detection in manga.

\section*{Acknowledgement}
This work was supported by JSPS KAKENHI Grant Number 23K16896 and the Kayamori Foundation of Informational Science Advancement.

\bibliographystyle{ACM-Reference-Format}
\bibliography{main}

\end{document}